\documentclass[sigconf]{acmart}
\AtBeginDocument{%
  }

\copyrightyear{2026}
\acmYear{2026}
\setcopyright{cc}
\setcctype{by}
\acmConference[SIGIR '26]{Proceedings of the 49th International ACM SIGIR Conference on Research and Development in Information Retrieval}{July 20--24, 2026}{Melbourne, VIC, Australia}
\acmBooktitle{Proceedings of the 49th International ACM SIGIR Conference on Research and Development in Information Retrieval (SIGIR '26), July 20--24, 2026, Melbourne, VIC, Australia}
\acmDOI{10.1145/3805712.3808604}
\acmISBN{979-8-4007-2599-9/2026/07}

\usepackage[most]{tcolorbox}
\usepackage{enumitem}
\usepackage{booktabs}
\usepackage{xcolor}
\usepackage{multirow}
\begin{document}

\title{NanoKnow: How to Know What Your Language Model Knows}

\author{Lingwei Gu}
\orcid{0009-0000-7598-1200}
\authornote{Equal Contribution}
\affiliation{
    \institution{University of Waterloo}
    \city{Waterloo}
    \state{ON}
    \country{Canada}}

\email{lingwei.gu@uwaterloo.ca}

\author{Nour Jedidi}
\orcid{0009-0007-0189-9678}
\authornotemark[1]
\affiliation{
    \institution{University of Waterloo}
    \city{Waterloo}
    \state{ON}
    \country{Canada}}

\email{njedidi@uwaterloo.ca}

\author{Jimmy Lin}
\orcid{0000-0002-0661-7189}
\affiliation{
    \institution{University of Waterloo}
    \city{Waterloo}
    \state{ON}
    \country{Canada}
}
\email{jimmylin@uwaterloo.ca}

\begin{abstract}
\emph{How do large language models (LLMs) know what they know}? Answering this question has been difficult because pre-training data is often a ``black box'' -- unknown or inaccessible. The recent release of nanochat -- a family of small LLMs with fully open pre-training data -- addresses this as it provides a transparent view into where a model's parametric knowledge comes from. Towards the goal of understanding how knowledge is encoded by LLMs, we release \emph{NanoKnow}, a benchmark dataset that partitions questions from Natural Questions and SQuAD into splits based on whether their answers are present in nanochat's pre-training corpus. Using these splits, we can now properly disentangle the sources of knowledge that LLMs rely on when producing an output. To demonstrate NanoKnow’s utility, we conduct experiments using eight nanochat checkpoints. Our findings show: (1) closed-book accuracy is strongly influenced by answer frequency in the pre-training data, (2) providing external evidence can mitigate this frequency dependence, (3) even with external evidence, models are more accurate when answers were seen during pre-training, demonstrating that parametric and external knowledge are complementary, and (4) non-relevant information is harmful, with accuracy decreasing based on both the position and the number of non-relevant contexts. We release all NanoKnow artifacts at \url{https://github.com/castorini/NanoKnow}. 
\end{abstract}



\maketitle

\begin{figure*}[t!]
  \centering
  \includegraphics[width=0.7\textwidth]{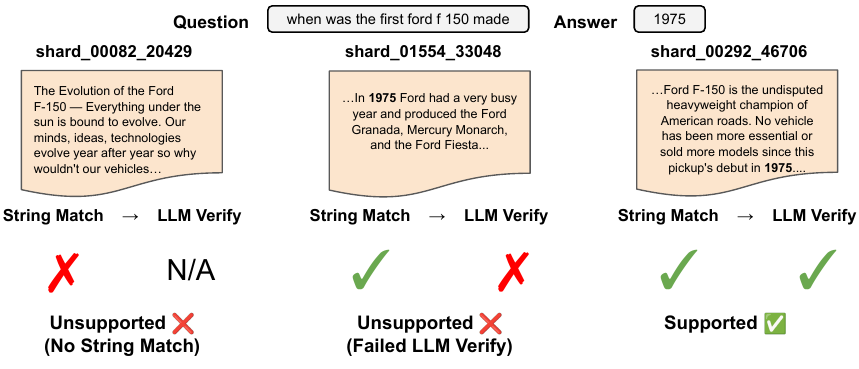}
  \caption{An example of NanoKnow on a question-answer pair. If any passage is deemed to answer the question after the string match and LLM verify steps, we label the question as ``supported''. Otherwise, the question is considered ``unsupported''.}
  \label{fig:methods}
\end{figure*}

\section{Introduction}

Large language models (LLMs) have demonstrated remarkable capabilities across a wide range of tasks,  yet it is unclear \emph{how they know what they know}. While LLMs ultimately express their knowledge through their outputs at inference time, \emph{how} and \emph{where} this knowledge is acquired remains an open question.  

Knowledge expressed by LLMs can originate from various, potentially entangled, sources. There exists knowledge stored within their parameters~\cite{roberts-etal-2020-much}, which can be probed via closed-book question answering~\cite{jiang2020can}, but this only tells us \emph{what} LLMs know, not necessarily \emph{how} that knowledge was acquired. For example, did this knowledge come from memorization of its pre-training data~\cite{carlini2022quantifying} or is the model performing a sort of multi-hop reasoning over  facts encoded within its parameters~\cite{yang2024large}? Alternatively, external knowledge can be injected into the LLM using retrieval-augmented generation (RAG), but in this case does the model's output solely represent facts present in the external context or does the output represent a latent interaction between the external context and the LLM's parametric knowledge~\cite{zhao2025understanding}? Ultimately, answering these questions requires understanding the model's pre-training data, but this has been difficult as 
such data is often unknown or inaccessible~\cite{liu-etal-2025-olmotrace}. 

Recently, this changed with developments in fully open LLMs, making the understanding of the pre-training data now possible. A notable example is the release of nanochat~\cite{nanochat}, which, by being pre-trained on the open FineWeb-Edu corpus -- a collection of educational web content ~\cite{penedo2024the} -- provides a completely transparent and self-contained environment for tracing the information an LLM has \emph{seen}. Such transparency allows us to answer questions like: does seeing facts more often make it easier to recall? When does RAG actually make a difference?  However, transparency of data is only the first step. Before we can answer these questions systematically,  we require a resource which can not only identify questions an LLM has seen the answer to during pre-training but also questions beyond its knowledge. Such a resource is a necessary step toward properly disentangling  and understanding the various sources of knowledge LLMs rely on when producing their outputs. 

To address this, we release NanoKnow, a benchmark dataset 
of questions from Natural Questions (NQ) ~\cite{kwiatkowski-etal-2019-natural} and SQuAD~\cite{rajpurkar-etal-2016-squad} projected onto the FineWeb-Edu corpus. NanoKnow partitions each dataset into two splits --  ``supported'' (questions for which the answer exists in the pre-training data) and ``unsupported'' (questions for which the answer does not exist in the pre-training data) -- enabling a controlled evaluation of knowledge in LLMs, like nanochat, which were pre-trained entirely on FineWeb-Edu. To generate these relevance judgments, NanoKnow was built in three stages. In the first stage, we build a searchable BM25 index over the corpus using Anserini~\cite{yang2017anserini} and retrieve candidate documents for each question. Next, we check for exact match answer strings across the retrieved documents. In the last stage, we use LLM-based verification to filter out coincidental matches, keeping only documents that genuinely answer the questions.

With NanoKnow in hand, we run comprehensive experiments using eight nanochat checkpoints across three different model scales. Our various experiments demonstrate the value of NanoKnow as a tool for \emph{confidently} disentangling and evaluating the contributions of different knowledge sources underlying an LLM's outputs. Using NanoKnow, we were able to confirm and replicate a range of results across the literature~\cite{carlini2022quantifying, biderman2023pythia, cuconasu2024power, liu-etal-2024-lost, wanggeneralization, kandpal2023large}, highlighting its reliability:

\begin{itemize}[leftmargin=15pt]
    \item[{(1)}] Closed-book question answering effectiveness is highly related to answer frequency in the pre-training corpus. We found a clear increase in nanochat's accuracy when it has ``seen'' the answer more often.
    \item[{(2)}] Integrating external evidence mitigates this dependence on  ``memorization'', but even with external evidence, nanochat is more effective on questions with a higher answer frequency in the pre-training corpus.
    \item[{(3)}] Even when provided the oracle answer document, nanochat was more accurate on ``supported'' versus ``unsupported'' questions, demonstrating that parametric knowledge can complement external knowledge.
    \item[{(4)}] Despite nanochat having ``seen'' the answer to a question, it is negatively impacted by distractor documents (i.e., non-relevant documents). We found a clear decline in accuracy based on \emph{where} the answer document is positioned with respect to distractors as well as \emph{how many} distractors are present. 
\end{itemize}

\noindent
We hope NanoKnow provides a foundation for future explorations in  understanding \emph{how LLMs know what they know}.
\section{NanoKnow}
\label{sec:dataset_projection}

NanoKnow is a benchmark dataset that partitions NQ and SQuAD questions into \textit{supported} and \textit{unsupported} splits based on the presence of their answers within  FineWeb-Edu~\cite{penedo2024the}. In this section, we present NanoKnow and describe our process for producing it. 

\subsection{Projection Data}

\paragraph{Corpus} We build NanoKnow on the shuffled version of the FineWeb-Edu corpus released by ~\citet{nanochat}.\footnote{\url{https://huggingface.co/datasets/karpathy/fineweb-edu-100b-shuffle}} This variant of FineWeb-Edu comes as 1,823 parquet shards, each containing thousands of web documents. The total corpus size is about 171GB and contains 97,230,848 documents. FineWeb-Edu was chosen in our experiments primarily due to the recent release of nanochat~\cite{nanochat}, a family of small LLMs that utilized it for pre-training. 

To enable a searchable BM25 index for our projection pipeline -- which we describe in the next subsection -- we build a FineWeb-Edu inverted index using Anserini~\cite{yang2017anserini}. The full index is about 326GB.

\paragraph{Question-Answering Datasets} We project the following question-answering (QA) benchmarks onto FineWeb-Edu:

\begin{itemize}[leftmargin=15pt]
    \item \textbf{Natural Questions (NQ)}~\cite{kwiatkowski-etal-2019-natural}: Open-domain questions from Google search queries. We use the validation set (3,610 questions). Each question has one or more short answers.
    \item \textbf{SQuAD}~\cite{rajpurkar-etal-2016-squad}: Reading comprehension questions where answers are spans from Wikipedia passages. We use the validation set (10,570 questions).
\end{itemize}

\subsection{Projection Pipeline}
\label{sec:pipeline}
Given a question and its corresponding \emph{gold} answer, we project it onto FineWeb-Edu using a three-step process, which we describe in this subsection.  A high-level illustration of the NanoKnow pipeline for an example QA pair is shown in Figure~\ref{fig:methods}.

\paragraph{Step 1: BM25 Retrieval}
Using BM25, we first search the index to retrieve documents that may contain the answer. We retrieve the top-100 candidate documents, leveraging  Pyserini~\cite{lin2021pyserini} for retrieval.

\paragraph{Step 2: Answer String Matching (String Match in Figure~\ref{fig:methods})}
\label{sec:has_answer}
Next, we check whether any retrieved document contains the gold answer string associated with the query. The gold answers are taken from the official evaluation set of each benchmark. We lowercase everything and strip extra whitespace, then look for the answer as a substring. If it shows up, we flag the document as a \textit{candidate match}. This is fast, but returns many false positives. For example, for the question ``What is the best bakery in Paris?'' the word ``Paris'' might appear in a document about the song Paris and not Paris, France. Another step is needed to filter these out.

\begin{table*}[!t]
\centering
\caption{Examples of supported and unsupported NQ questions. Failed LLM Verify examples contain the gold answer string in FineWeb-Edu but do not directly answer the question.}
\label{tab:examples}
\resizebox{\textwidth}{!}{%
\begin{tabular}{p{2.4cm}|p{5cm}|p{2cm}|p{7.8cm}}
\toprule
\textbf{Type} & \textbf{Question} & \textbf{Answer} & \textbf{Evidence} \\
\midrule
\multirow{4}{*}{\shortstack[l]{Supported}} &
When was the last time anyone was on the moon? &
\multirow{2}{*}{December 1972} & 
``No one has walked on the Moon since \textbf{December 1972}.'' \\
\cline{2-4}
&
What is the main artery that takes blood from the heart to the body? &
\multirow{2}{*}{The aorta} &
``Arteries begin with the \textbf{aorta}, the large artery leaving the heart.'' \\
\midrule
\multirow{4}{*}{\shortstack[l]{Unsupported\\(Failed LLM Verify)}} &
Who won last year's NCAA women's basketball? &
\multirow{2}{*}{South Carolina} &
``The Wildcats were ranked 8th in the nation before a loss to \textbf{South Carolina} on October 4.'' \\
\cline{2-4}
&
Love Yourself by Justin Bieber is about who? &
\multirow{2}{*}{Rihanna} &
``Other Icelandic artists have followed in his footsteps, collaborating with international acts like \textbf{Rihanna} and Ed Sheeran.'' \\
\midrule
\multirow{3}{*}{\shortstack[l]{Unsupported\\(No String Match)}} 
&
Who sang I ran all the way home? &
The Impalas &
\multicolumn{1}{c}{--} \\ 
\cline{2-4}
& 
Who plays Gram on The Young and the Restless? &
\multirow{2}{*}{Max Shippee} &
\multicolumn{1}{c}{\multirow{2}{*}{--}} \\
\bottomrule
\end{tabular}
}
\end{table*}

\paragraph{Step 3: LLM Verification (LLM Verify in Figure~\ref{fig:methods})}
To address potential false positives from the output of the String Match step, we next leverage an LLM to separate real answer string matches from coincidental matches. In particular, for each candidate answer document provided by String Match (retrieved documents that contain the answer string), we extract a context window around where the answer appears: 256 words before the match and 256 words after. We send this context to Qwen3-8B~\cite{yang2025qwen3} with the question and ask it to classify whether the context answers the question.

\medskip
\noindent
After completing these three steps, NQ and SQuAD can be broken down into two splits:

\begin{itemize}[leftmargin=15pt]
    \item \textbf{Supported}: Questions where the answer appears in FineWeb-Edu in a relevant context. These are questions which passed both String Match (step 2) and LLM Verify (step 3).   
    \item \textbf{Unsupported}: Questions where the answer does not appear in any retrieved document or only shows up in unrelated contexts. These are questions which either have no string match (step 2) or failed  LLM Verify (step 3). 
\end{itemize}

\noindent
With the set of supported questions, we  can build the NanoKnow relevance judgments (qrels in TREC parlance), which link the question to its answer document in FineWeb-Edu. Each document gets a unique ID that encodes its location in the corpus. The format is \texttt{shard\_XXXXX\_YYYYY}, where \texttt{XXXXX} is the zero-padded shard number and \texttt{YYYYY} is the row offset within that shard. For example, \texttt{shard\_00151\_20323} refers to row 20,323 in shard 151. This encoding lets us trace any answer document back to its exact location in the FineWeb-Edu parquet files. For unsupported questions, as they do not contain answers in FineWeb-Edu, we simply provide a file with the question text. 

\subsection{Projection Results}

Now that we have projected NQ and SQuAD onto FineWeb-Edu, we examine the resulting projection. In Table~\ref{tab:examples}, we show examples of questions from NQ which were labeled by the projection pipeline as supported or unsupported. 

Table~\ref{tab:projection_stats} shows the projection results. The ``String Match \emph{only}'' column shows the percentage of questions which have its answer string in at least one retrieved document. ``String Match $\rightarrow$ LLM Verify'' is the percentage of questions which survive after the LLM Verify step.  For NQ, 73.9\% of questions have the answer string in a retrieved document; after LLM Verify, 66.2\% are confirmed to be supported. SQuAD is higher at 70.9\% of questions deemed supported, with the LLM Verify step removing 8.0\% of questions as coincidental string matches. This is not surprising as SQuAD answers come from Wikipedia, and FineWeb-Edu has a lot of Wikipedia content. 

Lastly, to validate that our unsupported labels are accurate, we prompt the official d32 nanochat checkpoint to answer the unsupported questions in a closed-book setting.\footnote{\texttt{karpathy/nanochat-d32}}  We found a closed-book exact match (EM) answering accuracy of 
0.8\% and 1.5\% for NQ and SQuAD. More details on the EM accuracy metric can be found in Section~\ref{sec:experimental_setup}. 

\begin{table}[t!]
\centering
\caption{Projection rates for NQ and SQuAD on FineWeb-Edu. The reported percentage represents how many questions are supported after  String Match only and after String Match followed by LLM Verify.}
\label{tab:projection_stats}
\begin{tabular}{lrr}
\toprule
\textbf{} & \textbf{NQ} & \textbf{SQuAD} \\
\midrule
Total QA Pairs & 3,610 & 10,570 \\
\midrule
String Match \emph{only} & 73.9\% & 78.9\% \\
String Match $\rightarrow$ LLM Verify (Supported) & 66.2\% & 70.9\% \\
\bottomrule
\end{tabular}
\end{table}

\paragraph{Released Artifacts}

As the result of this pipeline, we release the following to support reproducibility and future research:

\begin{itemize}[leftmargin=15pt]
    \item \textbf{Qrels}: Relevance judgments mapping supported questions from NQ and SQuAD to the answer documents in FineWeb-Edu. 
    \item \textbf{Unsupported Questions}: The set of unsupported questions from NQ and SQuAD whose answers are not in FineWeb-Edu.
    \item \textbf{Lucene Index}: Pre-built index over FineWeb-Edu (326GB).\footnote{https://huggingface.co/datasets/LingweiGu/NanoKnow-Fineweb-Edu-Index}
    \item \textbf{Evaluation Code}: Scripts to reproduce all experiments, including LLM-Judge prompts and evaluation metrics.
\end{itemize}

\noindent
The artifacts are available at \url{https://github.com/castorini/NanoKnow}.
\section{Experimental Setup}
\label{sec:experimental_setup}

NanoKnow now allows us to answer many interesting questions regarding how pre-training data shapes what knowledge LLMs rely on. We study a subset of these questions: 

\begin{itemize}[leftmargin=15pt]
    \item Does seeing the answer to a question more often in pre-training improve closed-book QA accuracy? What if we integrate external evidence?
    \item How does closed-book QA compare to open-book QA on supported questions?
    \item How does an LLM's open-book QA accuracy differ on supported versus unsupported questions?
    \item On supported questions, how do distractors (i.e., non-relevant information) influence an LLM's QA accuracy?
\end{itemize}

\noindent
To answer these questions, we make use of the nanochat~\cite{nanochat} family of models, which were entirely pre-trained on the FineWeb-Edu corpus discussed in Section~\ref{sec:dataset_projection}. We consider three nanochat model sizes: d20 ($\approx$ 561M parameters); d32 ($\approx$ 1.9B parameters); and d34 ($\approx$ 2.2B parameters). To ensure the robustness of our results to any variations in how models were trained, each evaluation is run with multiple open-source checkpoints for each model scale. 

Across experiments, we use the following three prompting setups. The first is  ``Closed-Book'', where nanochat is only prompted with the question. The second is ``w/ FineWeb Context'', where nanochat is ``reminded'' with the oracle answer passage from its pre-training data; and, lastly, for SQuAD, we also evaluate nanochat with the original context, ``w/ Original Context'', where nanochat is provided the original answer context from SQuAD.\footnote{As NQ is an open-domain QA task, there does not exist a singular, default ``original context'', thus for simplicity we do not consider it in our experiments.}  As FineWeb-Edu documents are very long, for these experiments, we only consider the surrounding context window of 200 words around the first matched answer (approximately 100 words before and 100 words after the answer). 

To evaluate the accuracy of responses generated by nanochat, we use two approaches. The first is exact match (EM), computed using the standard evaluation scripts commonly used by the community for these benchmarks. EM checks if any of the predefined correct answers exactly appear in the model's output. If there is a match, the answer is deemed correct; otherwise the answer is deemed incorrect. The other method we consider is an LLM-Judge, which given nanochat's output and the predefined correct answers, classifies nanochat's output as correct or not. For this, we leverage Qwen3-14B~\cite{yang2025qwen3}. 

\begin{figure}[t!]
\centering
\includegraphics[width=0.7\linewidth]{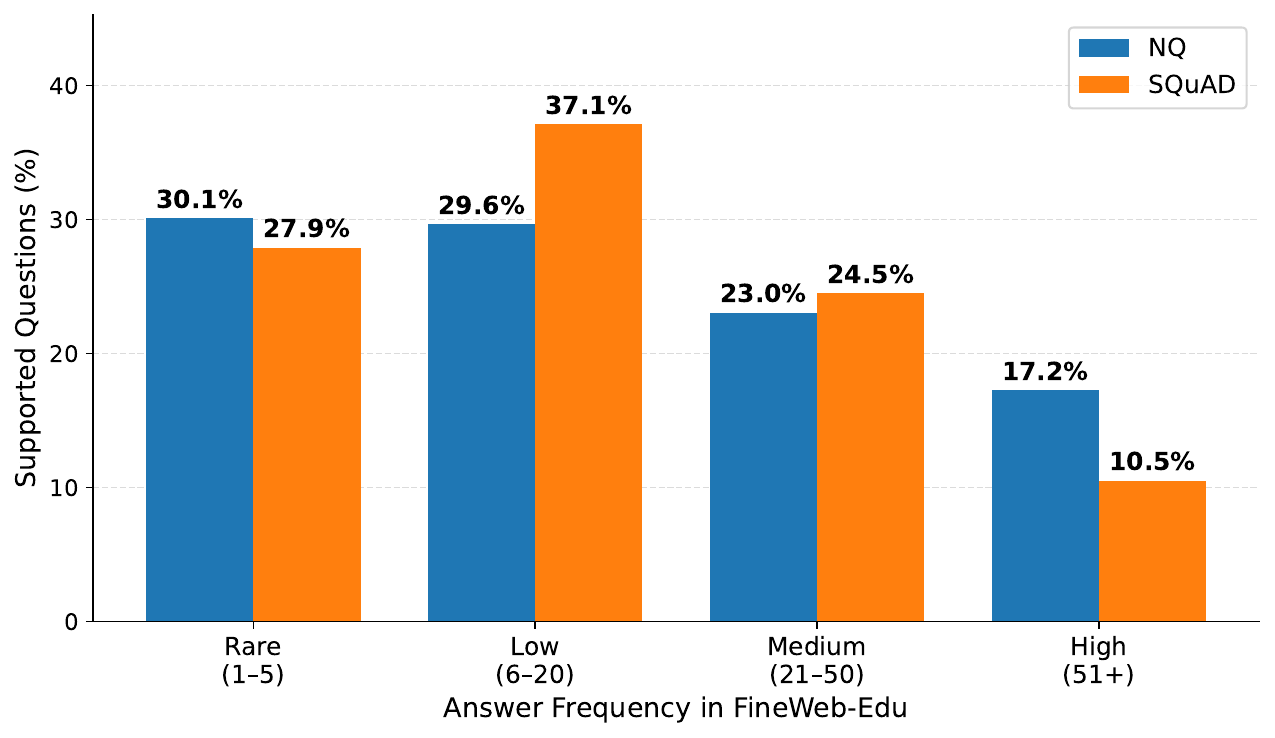}
\caption{Distribution of NanoKnow's supported questions by answer frequency in FineWeb-Edu for NQ and SQuAD.}
\label{fig:frequency_distribution}
\end{figure}

\begin{figure}[t]
\centering
\includegraphics[width=0.85\linewidth]{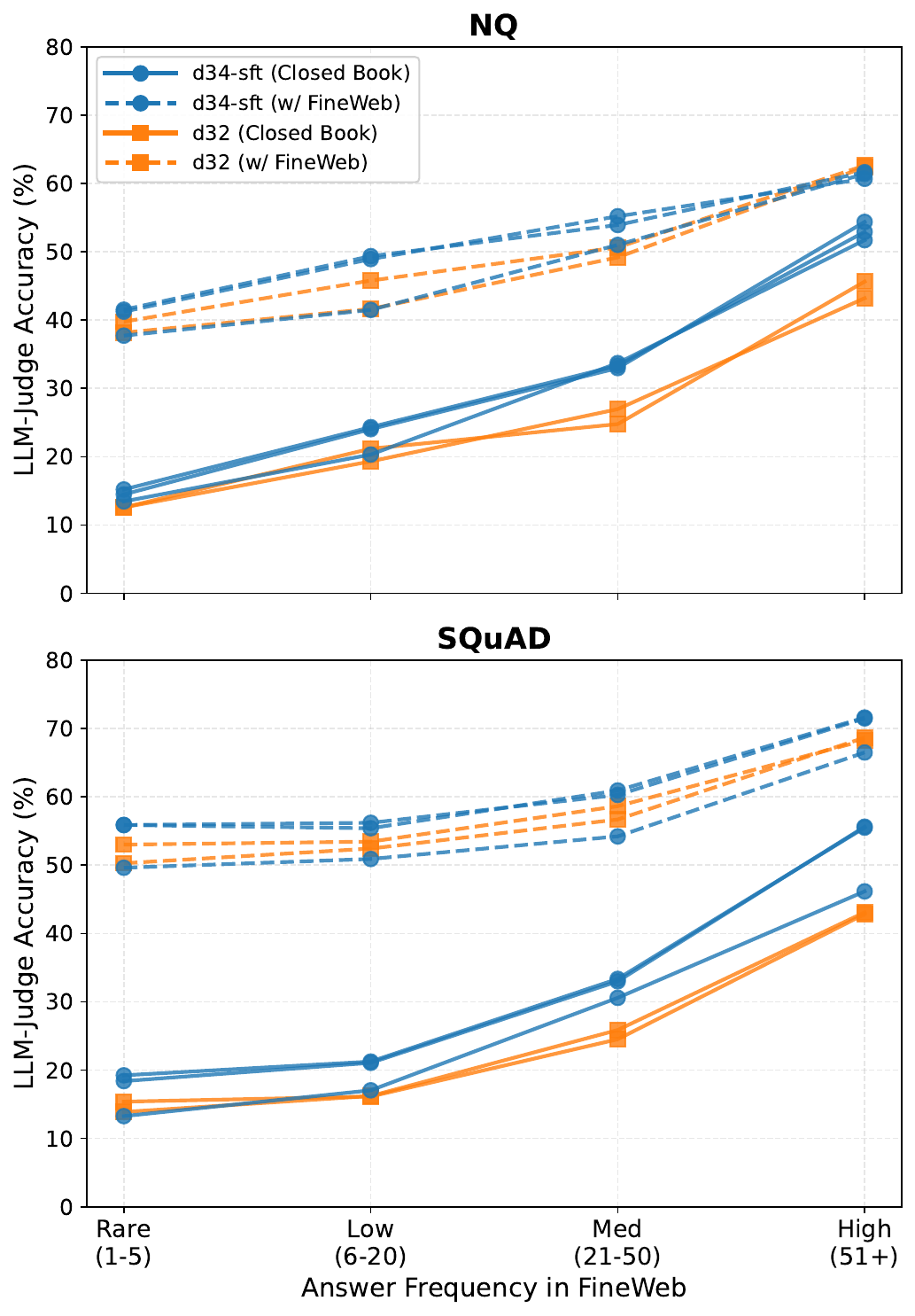}
\caption{Influence of pre-training data answer frequency on nanochat's accuracy. Solid lines show the closed-book prompt setup; dashed lines show w/ FineWeb-Edu context.}
\label{fig:answer_frequency_vs_accuracy}
\end{figure}

\begin{table*}[t]
  \centering
   \caption{Comparing closed-book versus open-book QA over the NQ and SQuAD supported splits of NanoKnow.}
   \resizebox{\textwidth}{!}{%
    \begin{tabular}{cl|c|cc|cc|cc|cc|cc}
    \toprule
    & \multicolumn{1}{c}{} & \multicolumn{1}{c|}{} &  \multicolumn{4}{c|}{\textbf{NQ}} & \multicolumn{6}{c}{\textbf{SQuAD}} \\
    \cmidrule(lr){4-7} \cmidrule(lr){8-13}
    & \multicolumn{1}{c}{} & \multicolumn{1}{c|}{} & \multicolumn{2}{c}{Closed-Book}
    & \multicolumn{2}{c|}{w/ FineWeb Context}
    & \multicolumn{2}{c}{Closed-Book}
    & \multicolumn{2}{c}{w/ FineWeb Context}
    & \multicolumn{2}{c}{w/ Original Context} \\
    \cmidrule(lr){4-5} \cmidrule(lr){6-7} \cmidrule(lr){8-9} \cmidrule(lr){10-11} \cmidrule(lr){12-13}
    &  \multicolumn{1}{l|}{\textbf{Model Checkpoint}} & \textbf{Model Size} &  EM & LLM-Judge & EM & LLM-Judge & EM & LLM-Judge & EM & LLM-Judge & EM & LLM-Judge \\
    \midrule
    1 & \texttt{sampathchanda/nanochat-d20} & \multirow{3}{*}{561M} & 0.004 & 0.008 & 0.022 & 0.018 & 0.004 & 0.002 & 0.021 & 0.019 & 0.086 & 0.081  \\
    2 & \texttt{shu127/nanochat-d20} &   & 0.003 & 0.005 & 0.021 & 0.016 & 0.003 & 0.011 & 0.016 & 0.013 & 0.042 & 0.040 \\
    3 & \texttt{pankajmathur/nanochat-d20} & & 0.003 & 0.014 & 0.028 & 0.023 & 0.005 & 0.008 & 0.022 & 0.020 & 0.056 & 0.052 \\
    \midrule
    4 & \texttt{karpathy/nanochat-d32} & \multirow{2}{*}{1.9B} & 0.196 & 0.224 & 0.468 & 0.476 & 0.114 & 0.173 & 0.465 & 0.540 & 0.672 & 0.736 \\
    5 & \texttt{Antigma/nanochat-d32} & & 0.198 & 0.226 & 0.516 & 0.492 & 0.122 & 0.169 & 0.483 & 0.551 & 0.686 & 0.740\\
    \midrule
    6 & \texttt{renatocastro33/nanochat-d34-sft} & \multirow{3}{*}{2.2B} & 0.250 & 0.283 & 0.503 & 0.528 & 0.167 & 0.228 & 0.512 & 0.587 & 0.721 & 0.779\\
    7 & \texttt{victoremnm/nanochat-d34-sft} & & 0.250 & 0.277 & 0.503 & 0.523 & 0.167 & 0.227 & 0.512 & 0.587 & 0.721 & 0.777\\
    8 & \texttt{pankajmathur/nanochat-d34-finetuned} & & 0.239 & 0.271 & 0.479 & 0.522 & 0.141 & 0.210 & 0.476 & 0.569 & 0.670 & 0.749\\
    \bottomrule
  \end{tabular} }
  \label{tab:SQuAD_nq_results}
\end{table*}

\begin{table}[t]  
  \centering
  \caption{QA accuracy for supported versus unsupported questions on SQuAD (w/ Original Context).}
  \resizebox{\columnwidth}{!}{%
  \begin{tabular}{l|c|cc|cc}
    \toprule
    \multicolumn{1}{c}{} & \multicolumn{1}{c|}{} & \multicolumn{2}{c|}{\textbf{Supported}} 
    & \multicolumn{2}{c}{\textbf{Unsupported}} \\
    \cmidrule(lr){3-4} \cmidrule(lr){5-6}
    \textbf{Model Checkpoint} & \textbf{Model Size}  & EM & LLM-Judge & EM & LLM-Judge \\
    \midrule
    \texttt{sampathchanda/nanochat-d20} & \multirow{3}{*}{561M} &  0.086 & 0.081 & 0.069 & 0.068 \\
    \texttt{shu127/nanochat-d20} & & 0.042 & 0.040 & 0.032 & 0.031 \\
    \texttt{pankajmathur/nanochat-d20} & & 0.056 & 0.052 & 0.051 & 0.054 \\
    \midrule
    \texttt{karpathy/nanochat-d32} & \multirow{2}{*}{1.9B} & 0.672 & 0.736 & 0.554 & 0.688 \\
    \texttt{Antigma/nanochat-d32} & & 0.686 & 0.740 & 0.553 & 0.680 \\
    \midrule
    \texttt{renatocastro33/nanochat-d34-sft} & \multirow{3}{*}{2.2B} &  0.721 & 0.779 & 0.610 & 0.737  \\
    \texttt{victoremnm/nanochat-d34-sft} & & 0.721 & 0.777 & 0.610 & 0.733  \\
    \texttt{pankajmathur/nanochat-d34-finetuned} & &  0.670 & 0.749 & 0.574 & 0.702  \\    
    \bottomrule
  \end{tabular} }
  \label{tab:supported_results}
\end{table}

\section{Results}

\subsection{Impact of Answer Frequency in Pre-training} 
\label{ref: answer_freq}
We begin by measuring how closed-book and open-book QA accuracy are impacted by how often the answer was replicated (i.e., ``seen'') during pre-training. To study this, we measure how accuracy changes with answer frequency in the pre-training corpus. 

To measure frequency,  for a given question, we count the number of FineWeb-Edu documents in which the answer was found and verified by the LLM in step 3 of Section~\ref{sec:pipeline}. We then categorize questions into four frequency buckets: Rare (1--5 verified documents), Low (6--20), Medium (21--50), and High (51+). Figure~\ref{fig:frequency_distribution} shows the distribution of supported questions across these buckets; the majority of questions fall in the Rare and Low frequency buckets for both NQ and SQuAD.

The relationship between answer frequency and nanochat's accuracy is shown in Figure ~\ref{fig:answer_frequency_vs_accuracy}. We find a clear increase in closed-book QA effectiveness on both NQ and SQuAD  as the answer frequency increases, with  accuracy more than doubling for questions with high answer frequency versus rare answer frequency. However, interestingly we did not find this to be the case for d20 (omitted from the plot), suggesting that, at smaller parameter counts, the LLM does not have the capacity to ``memorize'' information.

When integrating external evidence (i.e., open-book QA), there is also a general increase in accuracy as answer frequency in the pre-training data increases. However, the rate of this improvement, especially for SQuAD, is much lower than in the closed-book setting, demonstrating that RAG can help mitigate this dependence on pre-training frequency.

\subsection{Closed-Book QA vs. Open-Book QA} We next examine how much improvement external knowledge  provides over the LLM's parametric knowledge.  The results of this experiment can be found in Table ~\ref{tab:SQuAD_nq_results}. 

As expected, we see a clear upward trend in closed-book accuracy for both NQ and SQuAD as model size increases, demonstrating that larger nanochat checkpoints indeed memorize more of their training data~\cite{tirumala2022memorization, carlini2022quantifying}. In particular, for NQ,  the closed-book LLM-Judge accuracy improves by 26.9 points (19.2\%) when comparing  the best nanochat-d20 (row 3) to the best nanochat-d34 checkpoint (row 6). Similarly, with SQuAD,  accuracy improves by  21.7 points (19.7\%) when comparing the corresponding best checkpoints for nanochat-d20 (row 2) and nanochat-d34 (row 6). 

Comparing the ``Closed-Book'' versus ``w/ FineWeb Context'' columns, we find that all nanochat checkpoints see a large jump in accuracy when provided an answer passage from its pre-training data as additional context.  On average, the relative improvement of incorporating FineWeb context decreases as the nanochat model size increases. For example, on NQ, we see an average LLM-Judge accuracy improvement of 2.4$\times$, 2.1$\times$, and 1.9$\times$, for the d20, d32, and d34 model scales, respectively. With SQuAD, the average LLM-Judge accuracy improvement is 4.4$\times$, 3.2$\times$, and 2.6$\times$, for d20, d32, and d34. This result suggests that smaller models benefit more from open-book QA versus larger models. Furthermore, we find that each of the model checkpoints is more accurate on SQuAD when utilizing the original context versus the FineWeb context. This makes sense since the original context is tailored to answer the question; in other words, the FineWeb context is like a textbook, whereas the original context is the answer booklet. 

Finally, we examine how nanochat's open-book QA effectiveness differs on questions in which its answer appears in the pre-training corpus (supported) versus questions where it does not appear in the pre-training corpus (unsupported). For this experiment, we provide nanochat with the original context from SQuAD as the FineWeb context does not exist for the unsupported split.  The results are presented  in Table~\ref{tab:supported_results} and show that  nanochat is consistently more accurate on supported questions across all model scales. 

\begin{table*}[t]
  \centering
  \caption{Influence of distractors on nanochat's (\texttt{Antigma/nanochat-d32}) effectiveness on supported questions. A denotes the answer document, D denotes the distractor document, and Q denotes the question.}
  \resizebox{\textwidth}{!}{%
\begin{tabular}{cl|cccc|cccc|cccc}
\toprule
& \multicolumn{1}{c|}{} &  
\multicolumn{4}{c|}{\textbf{Far: [A, D, Q]}} 
& \multicolumn{4}{c|}{\textbf{Mid: [D, A, D, Q]}} 
& \multicolumn{4}{c}{\textbf{Near: [D, A, Q]}} \\
\cmidrule(lr){3-6} \cmidrule(lr){7-10} \cmidrule(lr){11-14}
& \multicolumn{1}{c|}{}
& \multicolumn{2}{c}{NQ} & \multicolumn{2}{c|}{SQuAD}
& \multicolumn{2}{c}{NQ} & \multicolumn{2}{c|}{SQuAD}
& \multicolumn{2}{c}{NQ} & \multicolumn{2}{c}{SQuAD} \\
\cmidrule(lr){3-4} \cmidrule(lr){5-6} \cmidrule(lr){7-8} \cmidrule(lr){9-10} \cmidrule(lr){11-12} \cmidrule(lr){13-14}
& \textbf{Setting}
& EM & LLM-Judge & EM & LLM-Judge
& EM & LLM-Judge & EM & LLM-Judge
& EM & LLM-Judge & EM & LLM-Judge \\
\midrule
    1 & Closed-Book & 0.198 & 0.226 & 0.122 & 0.169 & 0.198 & 0.226 & 0.122 & 0.169 & 0.198 & 0.226 & 0.122 & 0.169 \\
    2 & Distractor \emph{only} & 0.152 & 0.194 & 0.091 & 0.154  & 0.152 & 0.194 & 0.091 & 0.154 & 0.152 & 0.194 & 0.091 & 0.154 \\
    3 & Answer \emph{only} & 0.516 & 0.492 & 0.483 & 0.551 & 0.516 & 0.492 & 0.483 & 0.551 & 0.516 & 0.492 & 0.483 & 0.551 \\
    \midrule
    4 & Answer + 1 Distractor & 0.452 & 0.447 & 0.411 & 0.478 & \multicolumn{4}{c}{N/A} & 0.456 & 0.457 & 0.428 & 0.501 \\
    5 & Answer + 2 Distractors & 0.414 & 0.422 & 0.363 & 0.438 & 0.387 & 0.406 & 0.352 & 0.428 &  0.433 & 0.448 & 0.397 & 0.480 \\
    6 & Answer + 4 Distractors &  0.357 & 0.378 & 0.287 & 0.367 & 0.334 & 0.368 & 0.277 & 0.363 & 0.417 & 0.432 & 0.369 & 0.457 \\
    \bottomrule
  \end{tabular}}
  \label{tab:distractors}
\end{table*}

\subsection{Influence of Distractors}

Lastly, we are interested in understanding how nanochat is influenced by distractors (i.e., non-relevant contexts). For this experiment, we follow the setup in ~\citet{cuconasu2024power}. We prompt nanochat using three different placements of the answer contexts and the distractor: ``Far'' in which the answer context is placed furthest away from the question; ``Mid'', in which the answer context is placed in the middle of the prompt, in between different distractor contexts; and ``Near'', in which the answer context is placed closest to the question. We additionally consider a setting in which nanochat is only prompted with a distractor context (``Distractor \emph{only}''). In our setup, the distractor context was the highest ranked BM25 document which \emph{did not} contain the answer; when two distractors are used, i.e., for the ``Mid'' case, we took the top-2 such documents. To ensure that the distractor contexts have the same length as the answer contexts, we extracted a random 200-word snippet from each distractor document. 
 
We compare all distractor setups to the closed-book and w/ FineWeb context (``Answer \emph{only}'') settings shown in Table~\ref{tab:SQuAD_nq_results}. For this experiment, we focus on \texttt{Antigma/nanochat-d32}, the strongest d32 checkpoint. These experiments were run over the supported splits of NQ and SQuAD. The results are shown in Table ~\ref{tab:distractors}. 

Beginning by comparing closed-book (row 1) to the distractor only setting (row 2), we find that prompting nanochat with a non-relevant context does worse than utilizing the model's parametric knowledge, with the LLM-Judge accuracy dropping 3.2 and 1.5 points on NQ and SQuAD, respectively. 

The negative influence of the  distractor context on nanochat's effectiveness is further confirmed when comparing the answer only setting (row 3) to each of the answer + distractor settings (rows 4 to 6). When prompted with answer and distractor documents, as might be expected in a practical RAG setting, 
nanochat is consistently less accurate than when only prompted with the correct answer context, across all prompt setups (Far, Mid, Near). Furthermore, nanochat is also less accurate when prompted with more distractors. Using the ``Far'' prompt on NQ as a representative case, the LLM-Judge accuracy drops from 0.447 (1 distractor) to 0.378 (4 distractors). 

Lastly, the results show that nanochat is most accurate when the answer context is closest to the question. But notably, being closer to the question is only helpful when there are no distractors between the answer and the question, as nanochat has a ``lost in the middle'' effect~\cite{liu-etal-2024-lost}, where it is least effective when the answer context is placed between distractors. 
\section{Related Work}

\paragraph{Tracing an LLM's capabilities to its pre-training data} LLMs pick up a wide range of factual knowledge during their pre-training, but identifying \emph{where} in the pre-training data the LLM learned that knowledge remains an open research question. Much of the research in this area fits directly under the umbrella of training data attribution methods~\cite{akyurek2022towards, changscalable, sun2025enhancing}, which try to find the pre-training data that can explain a model's output. This has been commonly done via gradient-based or representation-based methods~\cite{sun2025enhancing}. Other works, such as FASTTRACK~\cite{chen-etal-2024-fasttrack}  and OLMoTrace~\cite{ liu-etal-2025-olmotrace} take a more retrieval-oriented approach, leveraging semantic clustering or lexical overlap to match the LLMs' outputs to training examples. In fact, it was shown by ~\citet{akyurek2022towards} that even BM25 can serve as a strong baseline for tracing a model’s outputs back to training examples.

There have also been other works that have proposed methods for mapping task-specific data back to pre-training data a priori, with the goal of understanding how knowledge contained in, or properties of, the pre-training data link to capabilities of LLMs on specific downstream tasks. For example, ~\citet{kandpal2023large} proposed an approach which counts how often specific question and answer entities appear in documents in the pre-training corpus, with the aim of studying how the number of relevant documents to a question relates to its answering accuracy -- similar to our experiment in Section \ref{ref: answer_freq}. More recently, ~\citet{wanggeneralization}, proposed a method to measure an LLM's ``memorization'' versus ``generalization'' by mapping its output distribution to the task-specific pre-training data frequency.

\paragraph{Interplay of parametric versus external knowledge} It has been shown in previous works~\cite{kandpal2023large} -- and further demonstrated in our experiments -- that LLMs struggle to recall knowledge which is less frequent in its pre-training data. Even more so, parametric knowledge can become obsolete as time goes on. To address these shortcomings,  RAG has been proposed as an approach to feed the LLM \emph{external} knowledge at inference time to guide its generated output. With this, various research~\cite{mallen2023not, farahani-johansson-2024-deciphering, qianmerge, xie2023adaptive} has focused on understanding this interplay between parametric knowledge within the LLMs and external knowledge provided to the LLM; see ~\citet{xu-etal-2024-knowledge-conflicts} for a survey on the topic. 

We note, however, that a large chunk of these works has focused on evaluating this interplay for LLMs in which the pre-training data is \emph{unknown}. This makes it difficult to properly disentangle the effects of parametric and external knowledge since the knowledge the LLM knows is unclear. NanoKnow provides a benchmark in which the interplay of the different knowledge sources can be explored confidently.  
\section{Conclusion}

In this paper, we set out to answer a simple question: \emph{how do LLMs know what they know?} Towards this goal, we introduced NanoKnow, a benchmark dataset that identifies questions which nanochat -- or any LLM entirely pre-trained on FineWeb-Edu -- has ``seen'' the answer to during pre-training, along with the pipeline to produce it. By projecting Natural Questions and SQuAD onto FineWeb-Edu, we found that over 66\% and 71\% of questions, respectively, have verifiable supported answers. NanoKnow now allows us to answer many interesting questions regarding how pre-training data shapes what language models know.

Using NanoKnow, we ran controlled experiments across various nanochat checkpoints, providing a clear picture of how parametric knowledge and external knowledge interact. What a model knows is largely a function of frequency: answers seen often are recalled reliably, while rare answers require external help. RAG closes this gap, improving accuracy where parametric knowledge is weakest. Even with RAG, models perform better on questions they have already seen, showing that parametric knowledge and external knowledge complement each other. At the same time, external knowledge is fragile: distractors degrade accuracy, particularly when the answer is buried in the middle of the context, highlighting the importance of retrieval precision in practical RAG systems. Taken together, these findings replicate a wide range of results in the literature, further underscoring the reliability of NanoKnow. 

While we built NanoKnow on FineWeb-Edu, the methodology used to create it can be extended to any open corpus. Indexing is a one-time cost, and projecting new benchmarks afterward is fast. This opens the door to questions we have not yet explored thoroughly: how does the topical composition of pre-training data influence downstream capabilities? Can answer frequency information guide more effective data curation? We leave these directions to future work and release NanoKnow for the community to conduct fair and controlled studies of how training data shapes what LLMs can and cannot do.

\section*{Acknowledgments}

This research was supported in part by the Natural Sciences and Engineering Research Council (NSERC) of Canada. Additional funding was provided by Snowflake and the Institute of Information \& Communications Technology Planning \& Evaluation (IITP) grant funded by the Korean Government (MSIT) (No.\ RS-2024-00457882, National AI Research Lab Project).
\bibliographystyle{ACM-Reference-Format}
\balance
\bibliography{main}

@inproceedings{lin2021pyserini,
  title={Pyserini: A Python Toolkit for Reproducible Information Retrieval Research with Sparse and Dense Representations},
  author={Lin, Jimmy and Ma, Xueguang and Lin, Sheng-Chieh and Yang, Jheng-Hong and Pradeep, Ronak and Nogueira, Rodrigo},
  booktitle={Proceedings of the 44th International ACM SIGIR Conference on Research and Development in Information Retrieval (SIGIR 2021)},
  pages={2356-2362},
  year={2021}
}

@inproceedings{yang2017anserini,
  title={Anserini: Enabling the Use of Lucene for Information Retrieval Research},
  author={Yang, Peilin and Fang, Hui and Lin, Jimmy},
  booktitle={Proceedings of the 40th International ACM SIGIR Conference on Research and Development in Information Retrieval (SIGIR 2017)},
  pages={1253--1256},
  year={2017}
}

@article{yang2025qwen3,
  title={{Qwen3 Technical Report}},
  author={Yang, An and Li, Anfeng and Yang, Baosong and Zhang, Beichen and Hui, Binyuan and Zheng, Bo and Yu, Bowen and Gao, Chang and Huang, Chengen and Lv, Chenxu and others},
  journal={arXiv preprint arXiv:2505.09388},
  year={2025}
}

@inproceedings{cuconasu2024power,
  title={The Power of Noise: Redefining Retrieval for RAG Systems},
  author={Cuconasu, Florin and Trappolini, Giovanni and Siciliano, Federico and Filice, Simone and Campagnano, Cesare and Maarek, Yoelle and Tonellotto, Nicola and Silvestri, Fabrizio},
  booktitle={Proceedings of the 47th International ACM SIGIR Conference on Research and Development in Information Retrieval (SIGIR 2024)},
  pages={719--729},
  year={2024}
}

@misc{nanochat,
  author = {Andrej Karpathy},
  title = {nanochat: The Best ChatGPT That \$100 Can Buy},
  year = {2025},
  publisher = {GitHub},
  url = {https://github.com/karpathy/nanochat}
}

@inproceedings{penedo2024the,
author = {Penedo, Guilherme and Kydl\'{\i}\v{c}ek, Hynek and Allal, Loubna Ben and Lozhkov, Anton and Mitchell, Margaret and Raffel, Colin and Von Werra, Leandro and Wolf, Thomas},
title = {The FineWeb Datasets: Decanting the Web for the Finest Text Data at Scale},
year = {2024},
booktitle = {Proceedings of the 38th International Conference on Neural Information Processing Systems},
}

@inproceedings{liu-etal-2025-olmotrace,
    title = "{OLM}o{T}race: Tracing Language Model Outputs Back to Trillions of Training Tokens",
    author = "Liu, Jiacheng  and
      Blanton, Taylor  and
      Elazar, Yanai  and
      Min, Sewon  and
      Chen, Yen-Sung  and
      Chheda-Kothary, Arnavi  and
      Tran, Huy  and
      Bischoff, Byron  and
      Marsh, Eric  and
      Schmitz, Michael  and
      Trier, Cassidy  and
      Sarnat, Aaron  and
      James, Jenna  and
      Borchardt, Jon  and
      Kuehl, Bailey  and
      Cheng, Evie Yu-Yen  and
      Farley, Karen  and
      Anderson, Taira  and
      Albright, David  and
      Schoenick, Carissa  and
      Soldaini, Luca  and
      Groeneveld, Dirk  and
      Pang, Rock Yuren  and
      Koh, Pang Wei  and
      Smith, Noah A.  and
      Lebrecht, Sophie  and
      Choi, Yejin  and
      Hajishirzi, Hannaneh  and
      Farhadi, Ali  and
      Dodge, Jesse",
    booktitle = "Proceedings of the 63rd Annual Meeting of the Association for Computational Linguistics (Volume 3: System Demonstrations)",
    year = "2025",
    pages = "178--188",
}

@inproceedings{tirumala2022memorization,
author = {Tirumala, Kushal and Markosyan, Aram H. and Zettlemoyer, Luke and Aghajanyan, Armen},
title = {Memorization Without Overfitting:  Analyzing the Training Dynamics of Large Language Models},
year = {2022},
booktitle = {Proceedings of the 36th International Conference on Neural Information Processing Systems},
}

@inproceedings{carlini2022quantifying,
  title={Quantifying Memorization Across Neural Language Models},
  author={Carlini, Nicholas and Ippolito, Daphne and Jagielski, Matthew and Lee, Katherine and Tramer, Florian and Zhang, Chiyuan},
  booktitle={The Eleventh International Conference on Learning Representations},
  year={2022}
}

@inproceedings{rajpurkar-etal-2016-squad,
    title = "{SQ}u{AD}: 100,000+ Questions for Machine Comprehension of Text",
    author = "Rajpurkar, Pranav  and
      Zhang, Jian  and
      Lopyrev, Konstantin  and
      Liang, Percy",
    booktitle = "Proceedings of the 2016 Conference on Empirical Methods in Natural Language Processing (EMNLP)",
    year = "2016",
    pages = "2383--2392"
}

@article{kwiatkowski-etal-2019-natural,
    title = "Natural Questions: A Benchmark for Question Answering Research",
    author = "Kwiatkowski, Tom  and
      Palomaki, Jennimaria  and
      Redfield, Olivia  and
      Collins, Michael  and
      Parikh, Ankur  and
      Alberti, Chris  and
      Epstein, Danielle  and
      Polosukhin, Illia  and
      Devlin, Jacob  and
      Lee, Kenton  and
      Toutanova, Kristina  and
      Jones, Llion  and
      Kelcey, Matthew  and
      Chang, Ming-Wei  and
      Dai, Andrew M.  and
      Uszkoreit, Jakob  and
      Le, Quoc  and
      Petrov, Slav",
    journal = "Transactions of the Association for Computational Linguistics",
    volume = "7",
    year = "2019",
    pages = "452--466",
}

@article{jiang2020can,
  title={How Can We Know What Language Models Know?},
  author={Jiang, Zhengbao and Xu, Frank F. and Araki, Jun and Neubig, Graham},
  journal={Transactions of the Association for Computational Linguistics},
  volume={8},
  pages={423--438},
  year={2020}
}

@inproceedings{yang2024large,
  title={Do Large Language Models Latently Perform Multi-Hop Reasoning?},
  author={Yang, Sohee and Gribovskaya, Elena and Kassner, Nora and Geva, Mor and Riedel, Sebastian},
  booktitle={Proceedings of the 62nd Annual Meeting of the Association for Computational Linguistics (Volume 1: Long Papers)},
  pages={10210--10229},
  year={2024}
}

@inproceedings{zhao2025understanding,
  title={Understanding Parametric and Contextual Knowledge Reconciliation within Large Language Models},
  author={Zhao, Jun and Yang, Yongzhuo and Hu, Xiang and Tong, Jingqi and Lu, Yi and Wu, Wei and Gui, Tao and Zhang, Qi and Huang, Xuanjing},
  booktitle={Proceedings of the 39th International Conference on Neural Information Processing Systems},
  year={2025}
}

@article{liu-etal-2024-lost,
    title = "Lost in the Middle: How Language Models Use Long Contexts",
    author = "Liu, Nelson F.  and
      Lin, Kevin  and
      Hewitt, John  and
      Paranjape, Ashwin  and
      Bevilacqua, Michele  and
      Petroni, Fabio  and
      Liang, Percy",
    journal = "Transactions of the Association for Computational Linguistics",
    volume = "12",
    year = "2024",
    pages = "157--173",
}

@inproceedings{wanggeneralization,
title={Generalization v.s. Memorization: Tracing Language Models{\textquoteright} Capabilities Back to Pretraining Data},
author={Xinyi Wang and Antonis Antoniades and Yanai Elazar and Alfonso Amayuelas and Alon Albalak and Kexun Zhang and William Yang Wang},
booktitle={The Thirteenth International Conference on Learning Representations},
year={2025},
}

@inproceedings{roberts-etal-2020-much,
    title = "How Much Knowledge Can You Pack Into the Parameters of a Language Model?",
    author = "Roberts, Adam  and
      Raffel, Colin  and
      Shazeer, Noam",
    booktitle = "Proceedings of the 2020 Conference on Empirical Methods in Natural Language Processing (EMNLP)",
    year = "2020",
    pages = "5418--5426"
}

@inproceedings{biderman2023pythia,
  title={Pythia: A Suite for Analyzing Large Language Models Across Training and Scaling},
  author={Biderman, Stella and Schoelkopf, Hailey and Anthony, Quentin Gregory and Bradley, Herbie and O’Brien, Kyle and Hallahan, Eric and Khan, Mohammad Aflah and Purohit, Shivanshu and Prashanth, USVSN Sai and Raff, Edward and others},
  booktitle={International Conference on Machine Learning},
  pages={2397--2430},
  year={2023},
}

@inproceedings{mallen2023not,
  title={When Not to Trust Language Models: Investigating Effectiveness of Parametric and Non-Parametric Memories},
  author={Mallen, Alex and Asai, Akari and Zhong, Victor and Das, Rajarshi and Khashabi, Daniel and Hajishirzi, Hannaneh},
  booktitle={Proceedings of the 61st Annual Meeting of the Association for Computational Linguistics (Volume 1: Long Papers)},
  pages={9802--9822},
  year={2023}
}

@inproceedings{kandpal2023large,
  title={Large Language Models Struggle to Learn Long-Tail Knowledge},
  author={Kandpal, Nikhil and Deng, Haikang and Roberts, Adam and Wallace, Eric and Raffel, Colin},
  booktitle={International Conference on Machine Learning},
  pages={15696--15707},
  year={2023},
}

@article{sun2025enhancing,
  title={Enhancing Training Data Attribution with Representational Optimization},
  author={Sun, Weiwei and Liu, Haokun and Kandpal, Nikhil and Raffel, Colin and Yang, Yiming},
  journal={arXiv preprint arXiv:2505.18513},
  year={2025}
}

@inproceedings{akyurek2022towards,
  title={Towards Tracing Knowledge in Language Models Back to the Training Data},
  author={Aky{\"u}rek, Ekin and Bolukbasi, Tolga and Liu, Frederick and Xiong, Binbin and Tenney, Ian and Andreas, Jacob and Guu, Kelvin},
  booktitle={Findings of the Association for Computational Linguistics: EMNLP 2022},
  pages={2429--2446},
  year={2022}
}

@inproceedings{
changscalable,
title={Scalable Influence and Fact Tracing for Large Language Model Pretraining},
author={Tyler A. Chang and Dheeraj Rajagopal and Tolga Bolukbasi and Lucas Dixon and Ian Tenney},
booktitle={The Thirteenth International Conference on Learning Representations},
year={2025},
}

@inproceedings{farahani-johansson-2024-deciphering,
    title = "Deciphering the Interplay of Parametric and Non-Parametric Memory in Retrieval-Augmented Language Models",
    author = "Farahani, Mehrdad  and
      Johansson, Richard",
    booktitle = "Proceedings of the 2024 Conference on Empirical Methods in Natural Language Processing (EMNLP)",
    year = "2024",
    pages = "16966--16977",
}

@inproceedings{qianmerge,
  title={"Merge Conflicts!'" Exploring the Impacts of External Knowledge Distractors to Parametric Knowledge Graphs},
  author={Qian, Cheng and Zhao, Xinran and Wu, Tongshuang},
  booktitle={First Conference on Language Modeling},
  year={2024}
}

@inproceedings{xie2023adaptive,
  title={Adaptive Chameleon or Stubborn Sloth: Revealing the Behavior of Large Language Models in Knowledge Conflicts},
  author={Xie, Jian and Zhang, Kai and Chen, Jiangjie and Lou, Renze and Su, Yu},
  booktitle={The Twelfth International Conference on Learning Representations},
  year={2023}
}

@inproceedings{xu-etal-2024-knowledge-conflicts,
    title = "Knowledge Conflicts for {LLM}s: A Survey",
    author = "Xu, Rongwu  and
      Qi, Zehan  and
      Guo, Zhijiang  and
      Wang, Cunxiang  and
      Wang, Hongru  and
      Zhang, Yue  and
      Xu, Wei",
    booktitle = "Proceedings of the 2024 Conference on Empirical Methods in Natural Language Processing (EMNLP)",
    year = "2024",
    pages = "8541--8565",
}

@inproceedings{chen-etal-2024-fasttrack,
    title = "{FASTTRACK}: Reliable Fact Tracing via Clustering and {LLM}-Powered Evidence Validation",
    author = "Chen, Si  and
      Kang, Feiyang  and
      Yu, Ning  and
      Jia, Ruoxi",
    booktitle = "Findings of the Association for Computational Linguistics: EMNLP 2024",
    year = "2024",
    pages = "5821--5836",
}
\end{document}